\documentclass[12pt]{article}

\usepackage[margin=1in]{geometry}
\usepackage{amsmath,amssymb}
\usepackage{graphicx}
\usepackage{hyperref}
\usepackage{url}
\usepackage{booktabs}
\usepackage{listings}
\usepackage{xcolor}
\usepackage{caption}
\usepackage{subcaption}
\usepackage{float}
\usepackage{enumitem}
\usepackage{parskip}
\usepackage{titlesec}

\definecolor{codegray}{rgb}{0.95,0.95,0.95}
\definecolor{codecomment}{rgb}{0.4,0.4,0.4}
\definecolor{codekeyword}{rgb}{0.13,0.29,0.53}
\definecolor{codestring}{rgb}{0.31,0.60,0.02}

\lstdefinestyle{cppstyle}{
  backgroundcolor=\color{codegray},
  commentstyle=\color{codecomment}\itshape,
  keywordstyle=\color{codekeyword}\bfseries,
  stringstyle=\color{codestring},
  basicstyle=\ttfamily\footnotesize,
  breakatwhitespace=false,
  breaklines=true,
  captionpos=b,
  keepspaces=true,
  language=C++,
  showspaces=false,
  showstringspaces=false,
  showtabs=false,
  tabsize=2,
  frame=single,
  rulecolor=\color{gray!40}
}
\hypersetup{
  colorlinks=true,
  linkcolor=blue!70!black,
  citecolor=blue!70!black,
  urlcolor=blue!70!black,
}

\title{\textbf{On-Device Vision Training, Deployment, and Inference\\
on a Thumb-Sized Microcontroller}\\[0.4em]
{\large A Transparent, Single-File Foundation for Embedded Machine Learning}\\[4pt]
{\large on-device-vision-ai (Paper~1 of the webmcu-ai Series)}}

\author{Jeremy Ellis\\
High School Robotics Educator, British Columbia, Canada\\
\texttt{https://github.com/webmcu-ai}}

\date{April 2026}

\begin{document}

\maketitle

\begin{abstract}
This paper presents a complete, end-to-end on-device vision machine learning
pipeline---comprising data acquisition, two-layer CNN training with Adam
optimization, and real-time inference---executing entirely on a
microcontroller-class device costing \$15--40~USD.
Unlike cloud-based workflows that require external infrastructure and conceal
the computational pipeline from the practitioner, this system implements every
step of the core ML lifecycle in approximately 1,750~lines of readable C++ that
compiles in under one minute using the Arduino IDE, with no external ML
dependencies.
Running on the Seeed Studio ESP32-S3 XIAO ML Kit (8~MB PSRAM), the firmware
achieves three-class 64$\times$64 image classification in approximately
9~minutes per training run, with real-time inference at \textbf{6.3~FPS}.
Key contributions include: correct batch-level gradient accumulation;
pre-computed resize lookup tables for inference; dual-format weight export for
SD-free \emph{baked-in} deployment; a three-tier weight priority system
(SD binary $>$ baked-in header $>$ He-initialization) resolved automatically
at boot; a single-constant network reconfiguration interface; and
PSRAM-aware memory management suited to microcontroller constraints.
All source code and reference datasets are released under the MIT License at
\url{https://github.com/webmcu-ai/on-device-vision-ai}.
\end{abstract}

\textbf{Keywords:} embedded machine learning, ESP32-S3, microcontroller vision,
on-device training, TinyML, privacy-preserving ML, transparent ML pipeline,
engineering education, edge intelligence.

\tableofcontents
\newpage

\section{Introduction}
\label{sec:intro}

\subsection{The Case for On-Device Training}
\label{subsec:motivation}

Machine learning has become a foundational technology, yet its dominant workflow
remains stubbornly centralized: data leaves the device, travels to a cloud
datacenter, training occurs on remote hardware, and a pre-compiled model or library
is shipped back for local inference.
This pattern is so well established that most practitioners treat it as
inevitable---yet it carries costs that are frequently underappreciated in both
research and education.

\paragraph{Positioning relative to cloud-based platforms.}
Cloud-based TinyML platforms such as Edge Impulse provide powerful end-to-end
pipelines for dataset management, training, and deployment.
However, they are not required for all use cases.
This work demonstrates that, for small-scale and domain-specific problems, both
training and inference can be performed directly on low-cost microcontrollers,
making the approach viable for practical deployments where cloud
dependency, ongoing subscription costs, or data-privacy constraints are concerns.
The goal is not to replace cloud tools but to provide a complementary,
low-dependency alternative for rapid prototyping, small business deployment,
and educational use.

\paragraph{Privacy.}
When raw sensor data---images, audio, biometrics, medical readings, or industrial
signals---is uploaded to a cloud provider, the data owner surrenders physical
custody of it.
Even with contractual protections, the surface area for unauthorized access,
breach, or re-identification grows with every byte transmitted.
For medical devices, factory quality-control cameras, classroom learning tools, or
consumer IoT products, this is not a hypothetical concern.
On-device training keeps raw data local: only the model (a compact numerical
summary) ever leaves the hardware, and only when the owner chooses to share it.

\paragraph{Pipeline transparency.}
Production cloud frameworks based on TensorFlow and PyTorch are optimized for
throughput, not legibility.
High-level APIs abstract away backpropagation, gradient computation, and optimizer
state.
Energy consumed during training is invisible inside a remote datacenter.
Researchers who want to modify the learning algorithm---to test a new optimizer,
a novel regularization strategy, or an alternative activation function---must
navigate millions of lines of compiled framework code.
The entire computational pipeline is, in the words of practitioners, a
\emph{partial black box}~\cite{reddi2024mlsysbook}.

\paragraph{Complementary vendor frameworks.}
Espressif provides two official ML libraries for the ESP32-S3---ESP-DL and
ESP-NN---which are well-suited for deploying pre-trained, quantized production
models at high speed (see Section~\ref{sec:related}).
Neither supports on-device training.
The system described here is therefore complementary rather than competitive:
it occupies the training and transparency tier that vendor frameworks do not
address.

\paragraph{Educational need.}
Engineering students must ultimately understand \emph{when} to use a cloud
platform, \emph{when} to train on a local computer, and \emph{when}---now a
genuine option---to train fully on-device.
A curriculum focused solely on cloud-based workflows may limit opportunities for
students to engage with low-level ML system design.
Vijay Janapa Reddi observes that existing systems often optimize for performance over comprehension, noting that ``Our educational systems continue to separate the teaching of machine learning algorithms from the systems knowledge required to achieve computational scale''~\cite{reddi2024mlsysbook}.
Transparent, embedded implementations allow students to read every line of
backpropagation, add a debug print to any activation, and measure actual power
draw with an ammeter or a dedicated tool such as the Nordic PPK2---experiences
that are impossible with cloud training.
Following the pedagogical tradition of systems like TinyTorch, this implementation
is intended for first-principles comprehension: every layer, gradient, and
optimizer step is directly readable in the source, modifiable without a build
system, and observable live in the serial monitor.
Equally important: when a student later uses a cloud framework, they should
understand what it is doing on their behalf and why it is the correct tool for
that deployment context.
That understanding requires having built the simpler version first.

\paragraph{The WebSerial middle tier.}
Between the milliwatt constraints of a 240~MHz microcontroller and the
multi-TOPS capability of a cloud GPU cluster lies a middle tier that is only
beginning to be explored: the practitioner's own laptop, connected to an
embedded device via USB and the browser's WebSerial
API~\cite{webserial2024}.
WebSerial allows a web page to communicate with a serial port without installing
any driver software or Python toolchain, making it an ideal bridge for
resource-constrained learners and classroom settings.
Each paper in this series pairs a single-file on-device \texttt{.ino} with a
companion web application: \texttt{webmcu-vision-web} (Paper~2, vision),
\texttt{webmcu-audio-web} (Paper~3, audio), and \texttt{webmcu-motion-web}
(Paper~4, motion).
The present paper documents the on-device vision firmware that forms the
foundation of this approach.

\subsection{Three-Tier Mental Model for Learners}
\label{subsec:tiers}

This paper proposes that ML education should be structured around three deployment tiers,
each with a distinct privacy profile, infrastructure requirement, and appropriate
use case (Table~\ref{tab:tiers}).

\begin{table}[H]
\centering
\caption{Three-tier ML deployment model for engineering education.}
\label{tab:tiers}
\begin{tabular}{p{2.7cm} p{3.4cm} p{3.7cm} p{3.2cm}}
\toprule
\textbf{Tier} & \textbf{Infrastructure} & \textbf{When to choose} & \textbf{Privacy} \\
\midrule
\textbf{1. On-device}
  & Microcontroller only
  & Sensitive data, no connectivity, ultra-low power, transparent education
  & Raw data never leaves device \\
\addlinespace
\textbf{2. Local-computer} (WebSerial)
  & Laptop + USB cable; browser only
  & Larger datasets, interactive dashboards, no cloud cost, classroom use
  & Data stays on local machine \\
\addlinespace
\textbf{3. Cloud}
  & Remote GPU cluster
  & Production scale, large teams, foundation models, high-volume or multi-site inference
  & Data leaves premises \\
\bottomrule
\end{tabular}
\end{table}

This paper documents Tier~1;
the companion paper (Paper~2) will document Tier~2.

\subsection{Technical Contributions of This Work}
\label{subsec:contributions}
\begin{enumerate}
  \item \textbf{End-to-end on-device pipeline.}
        Data capture, CNN training with Adam optimizer, weight persistence, and
        real-time inference on a single \$15--40 microcontroller with no
        external computation.

  \item \textbf{6.3~FPS real-time inference.}
        Pre-computed resize lookup tables eliminate per-frame floating-point
        division, and per-frame pixel normalization replaces division by
        multiplication (\texttt{*~0.003921569f} in place of \texttt{/~255.0f}),
        achieving 159~ms per frame (6.3~FPS) in serial-output mode,
        with the OLED updated every 10th inference.

  \item \textbf{Correct batch gradient accumulation.}
        Weight gradient accumulators are zeroed once per batch boundary, not
        per-image, ensuring all images in a mini-batch contribute to each
        Adam update as intended.
        Per-image propagation buffers (\texttt{myDense\_grad},
        \texttt{myPool1\_grad}, \texttt{myConv1\_grad}) are zeroed per-image,
        as they carry the error signal \emph{through} the network rather than
        accumulating weight updates.

  \item \textbf{Dual-format weight export and baked-in deployment.}
        Trained weights are saved as a binary blob (\texttt{myWeights.bin}) and
        a C header (\texttt{myWeights.h}).
        Enabling \texttt{USE\_BAKED\_WEIGHTS} compiles weights into flash for
        SD-free inference, while the SD card remains available for further
        fine-tuning.

  \item \textbf{Three-tier weight priority system.}
        SD binary $>$ baked-in header $>$ random He-initialization, resolved
        automatically at boot with no user intervention.

  \item \textbf{Single-constant network and class reconfiguration.}
        \texttt{INPUT\_SIZE} scales all layer dimensions at compile time;
        \texttt{NUM\_CLASSES} and \texttt{myClassLabels[]} reconfigure the
        output layer, SD folder structure, OLED labels, and inference output.
        \begin{lstlisting}[language=C++]
#define NUM_CLASSES 3
String myClassLabels[NUM_CLASSES] = {"0Blank", "1Cup", "2Pen"};
#define INPUT_SIZE 64
        \end{lstlisting}

  \item \textbf{PSRAM-aware memory management.}
        All large tensors are allocated once at startup via \texttt{ps\_malloc()}
        and \texttt{ps\_calloc()}, with Adam moment buffers zero-initialized.
        A 172~KB RGB working buffer (\texttt{myRgbBuffer}) is pre-allocated
        globally and reused across inference, training image loads, and OLED
        rendering, avoiding repeated heap fragmentation from per-image
        allocation.
        Filter-stride offsets (e.g., \texttt{ob = f * OUTPUT\_SIZE *
        OUTPUT\_SIZE}) are hoisted outside inner loops to reduce repeated
        multiplication in the forward pass.

  \item \textbf{Configurable validation split.}
        A per-class hold-out set (\texttt{VALIDATION\_IMAGES}, default~3)
        provides a first generalization signal after each training run.
        \begin{lstlisting}[language=C++]
int VALIDATION_IMAGES = 3; // last N images per class held out (0 = disabled)
        \end{lstlisting}

  \item \textbf{Responsive training exit.}
        Touch and serial exit are checked every third training image rather
        than at end of batch, allowing the user to interrupt and save without
        waiting up to 15~seconds.

  \item \textbf{Transparent single-file implementation.}
        All logic resides in one $\sim$1{,}750-line Arduino sketch with no
        external ML dependencies, making the entire system forkable and
        curriculum-ready.
\end{enumerate}

\section{Related Work}
\label{sec:related}

\paragraph{ESP-DL and ESP-NN (Espressif)~\cite{espdl_github,espnn_github}.}
The official vendor inference libraries for the ESP32-S3.
ESP-NN provides hand-optimized SIMD kernels exploiting the chip's vector
instruction extensions; ESP-DL provides a higher-level model runner with INT8
quantized execution.
Both require a model to be trained externally in PyTorch or TensorFlow,
quantized via a separate toolchain, and converted to a proprietary binary format
before deployment to the device.
Neither supports on-device training of any kind, and neither exposes the training
pipeline for study or modification.
They represent the correct choice for production inference deployment on ESP32-S3
hardware; the present work is complementary.
To the best of the author's knowledge, and based on publicly available documentation at the
time of writing, no other system performs the core ML lifecycle entirely on the
ESP32-S3 without external computation.

\paragraph{TinyTorch.}
Vijay Janapa Reddi's TinyTorch~\cite{reddi2024tinytorch} demonstrates ML system
fundamentals through a transparent, software-only Python implementation.
This work extends those pedagogical principles to embedded hardware: a physical
camera, SD storage, a power supply, and a sub-watt compute budget.

\paragraph{MIT Tiny Training Engine.}
Lin et al.~\cite{lin2022ondevice} achieve on-device training under 256~KB using
quantization-aware sparse updates, focusing on memory minimization for production
deployment.
Their goal is orthogonal to ours: a practitioner benefits from understanding the
full-precision pipeline before studying quantization.

\paragraph{Fraunhofer AIfES.}
AIfES~\cite{wulfert2024aifes,aifes_github} is a professional embedded ML
framework supporting CNN training across diverse platforms in 50{,}000+ lines of
modular code.
Comprehensive, but requiring significant integration effort.
This work is a single-file, vision-specific application where every component is
immediately readable in the Arduino IDE.

\paragraph{TinyML4D.}
The TinyML4D initiative~\cite{plancher2024tinyml4d} scales embedded ML education
globally through open curricula.
Brian Plancher highlights the importance of the full life cycle, stating that
education should ``physically design, develop, deploy, and manage trained ML
models\ldots providing experience with the complete industrial ML
workflow''~\cite{plancher2024tinyml4d}.
Our system provides a concrete, low-cost artefact for such curricula
(note: Jeremy Ellis appears as a co-author in~\cite{plancher2024tinyml4d}).

\paragraph{TinyOL.}
Ren et al.~\cite{ren2021tinyol} present online learning on microcontrollers,
targeting continual adaptation; this system targets pedagogical completeness of
the initial training lifecycle.

\paragraph{Energy benchmarking.}
Kari\'{c} et al.~\cite{karic2025send} benchmark CNN inference energy against data
transmission energy in IoT, finding transmission costs can dominate.
The transparent power profiling in this work (Section~\ref{sec:energy}) makes the
\emph{training} energy directly measurable in a system of this kind.

\section{System Architecture}
\label{sec:architecture}

\subsection{Hardware Platform}
\label{subsec:hardware}

\begin{figure}[t]
\centering
\includegraphics[width=0.48\textwidth]{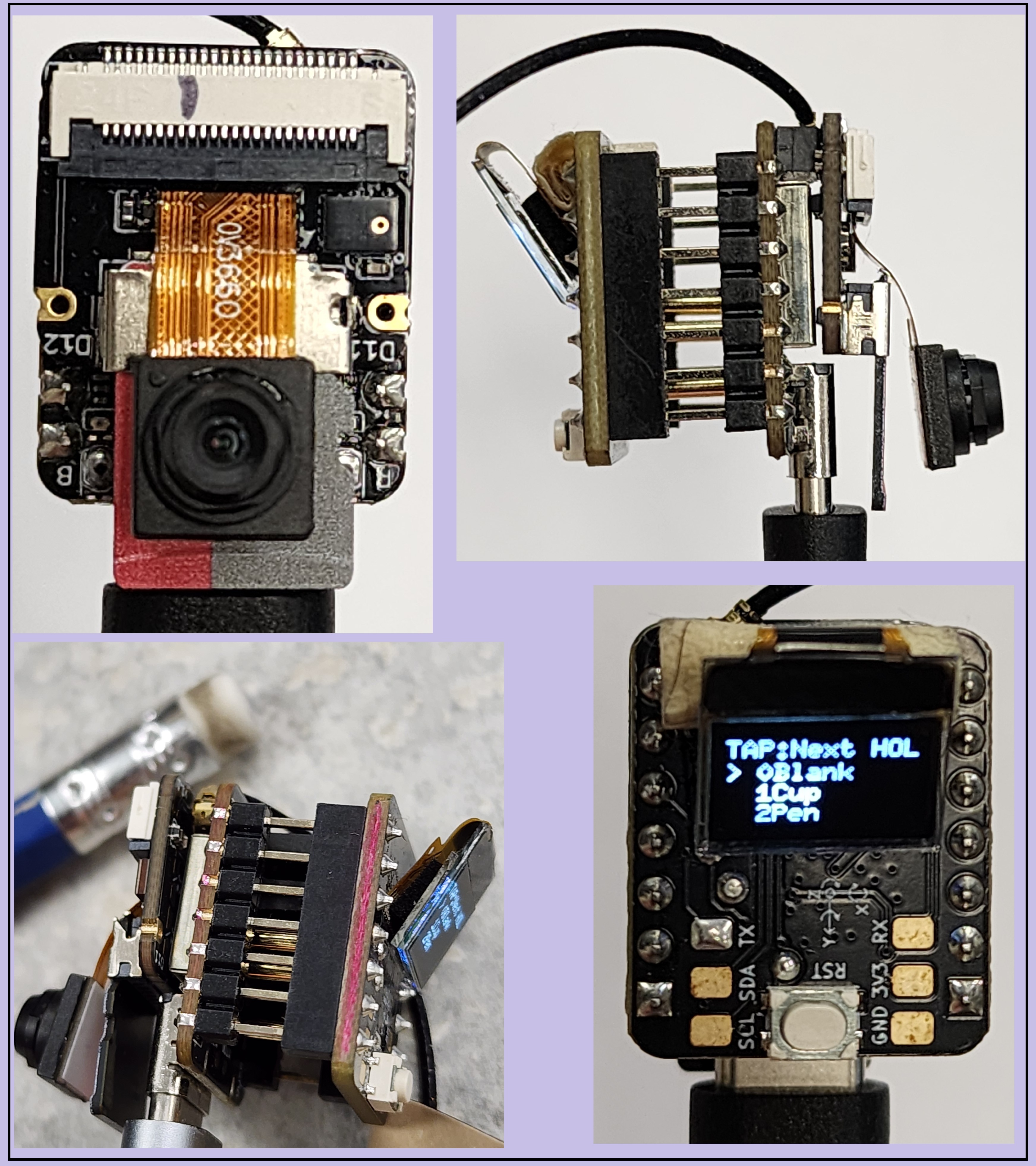}
\caption{XIAO ML Kit showing the ESP32-S3 board with integrated OV2640 camera,
72$\times$40 OLED display, and IMU sensor.
The compact form factor (8~MB PSRAM)
is sufficient for complete on-device training.
Pin~A0 capacitive touch enables
menu navigation via tap and long-press gestures.}
\label{fig:hardware}
\end{figure}

The system targets the Seeed Studio XIAO ESP32-S3 Sense board, available as a
standalone unit or as part of the XIAO ML Kit (Figure~\ref{fig:hardware}).

\begin{itemize}[nosep]
  \item ESP32-S3 dual-core Xtensa LX7 @ 240~MHz with hardware FPU
  \item 512~KB internal SRAM, \textbf{8~MB PSRAM} (OPI mode), 8~MB Flash
  \item OV2640 camera (up to 1600$\times$1200; used at 240$\times$240 JPEG)
  \item MicroSD slot (FAT32, up to 32~GB; required for training, optional after
        baked-in deployment)
  \item 72$\times$40 OLED display (SSD1306, via U8g2 library; optional but
        recommended)
  \item Multi-tap capacitive touch sensor on Pin~A0
  \item WiFi and BLE (present; unused in this work)
  \item Operating voltage: 3.3~V
\end{itemize}

The ML Kit retails for approximately \$22--40~USD depending on configuration;
the bare XIAO ESP32-S3 Sense board is available for under \$15~USD, operating
with serial-monitor output only.

\subsection{Repository Organization}
\label{subsec:repos}

The \textbf{webmcu-ai} GitHub organization (\url{https://github.com/webmcu-ai})
hosts the following repositories relevant to this paper:

\begin{description}[leftmargin=2em, style=nextline]
  \item[\texttt{on-device-vision-ai}]
    The on-device firmware described in this paper,
    organized as a flat structure to aid discoverability:
    \begin{itemize}[nosep]
      \item \texttt{firmware.ino} --- the complete Arduino sketch (v1.0.0):\\
            \url{https://github.com/webmcu-ai/on-device-vision-ai/blob/main/firmware.ino}
      \item \texttt{datasets/} --- three reference datasets of increasing size
            (Dataset~A: 2~classes~$\times$~9~images, smoke-test; Dataset~B:
            3~classes~$\times$~$\sim$30~images, matching the reported experiments;
            Dataset~C: 6~classes~$\times$~$\sim$60~images):\\
            \url{https://github.com/webmcu-ai/on-device-vision-ai/tree/main/datasets}
      \item \texttt{main.pdf} --- this paper;
            \texttt{CITATION.cff} --- citation metadata
    \end{itemize}
  \item[\texttt{webmcu-vision-web}]
    The companion browser-based WebSerial application, documented in Paper~2
    of this series (link to be added upon publication).
\end{description}

\subsection{Software Architecture}
\label{subsec:software}

The firmware is a single Arduino sketch of approximately 1,750~lines, organized
into five conceptually modular parts:

\begin{description}[leftmargin=2em, style=nextline]
  \item[Part~0: Core System]
    Preprocessor defines, global PSRAM buffer declarations, weight initialization
    priority logic, camera configuration (with horizontal mirror for natural
    front-facing orientation), SD mount with soft-failure timeout, OLED
    initialization, and the Arduino \texttt{setup()} / \texttt{loop()}.
  \item[Part~1: Data Collection]
    Live 240$\times$240 camera preview (black-and-white OLED rendering), touch-triggered
    JPEG capture, and class-labeled folder storage on SD.
    The camera buffer is drained every 50~ms to prevent frame-buffer overflow;
    the OLED is updated separately at 250~ms intervals to avoid unnecessary
    RGB conversion overhead.
    Display scale factors are computed once outside the pixel rendering loop.
  \item[Part~2: Training Engine]
    Forward propagation, backpropagation with correct batch-level gradient
    accumulation, Adam optimizer, epoch/batch loop, configurable validation
    evaluation, and dual-format weight export.
  \item[Part~3: Inference and Diagnostics]
    Optimized forward pass with pre-computed resize lookup tables and
    multiplication-based pixel normalization; 6.3~FPS in serial mode;
    black-and-white OLED live view with label overlay updated every 10th frame.
  \item[Part~4: Menu System]
    Multi-tap capacitive touch navigation with configurable thresholds and
    debounce;
    OLED menu rendering; serial command aliases (\texttt{t}/\texttt{L}).
\end{description}

\begin{figure}[t]
\centering
\includegraphics[width=\textwidth]{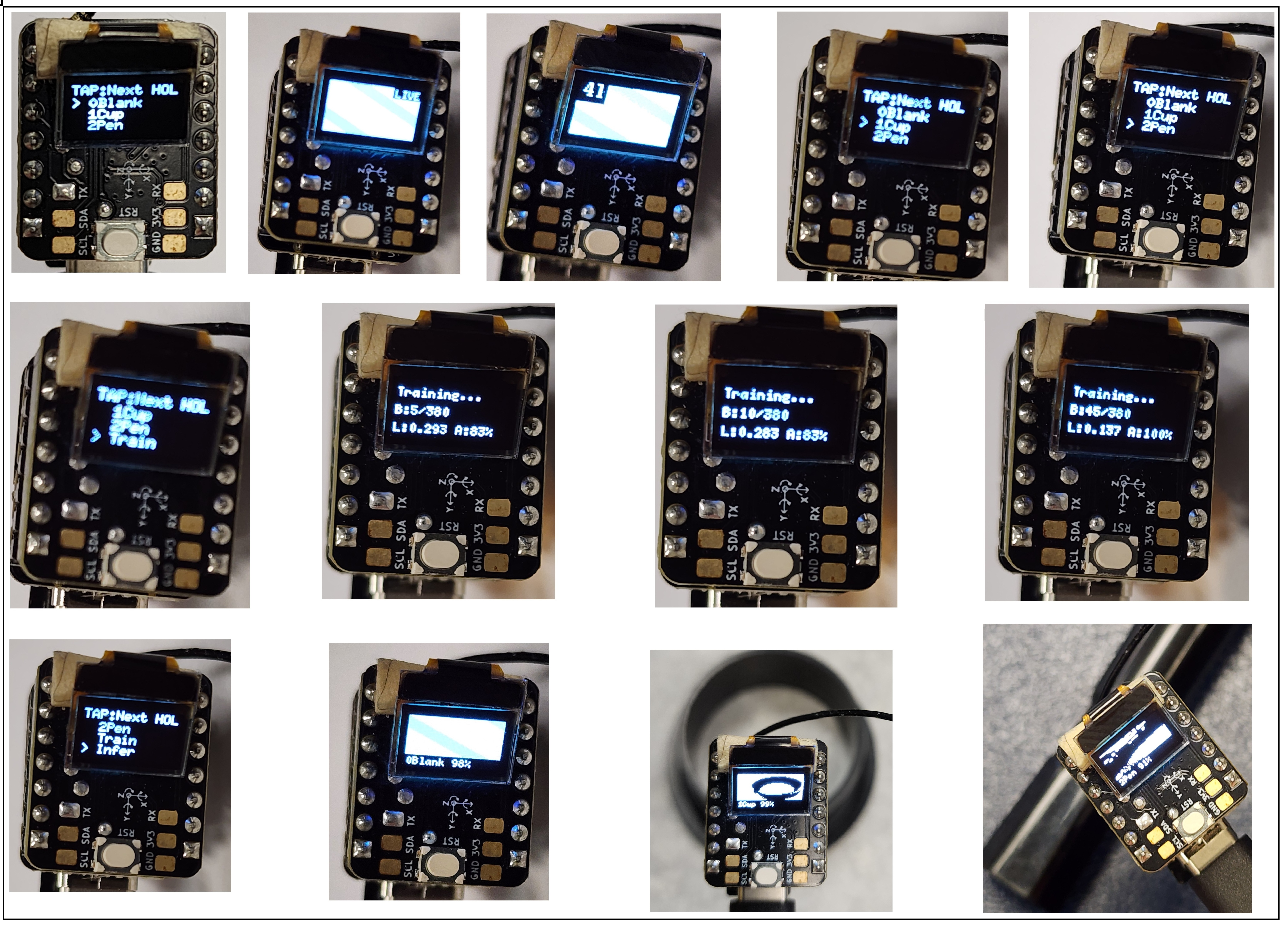}
\caption{OLED display showing the capacitive touch menu. Users navigate between
data collection, training, and inference modes using tap and long-press gestures
on Pin~A0.
Training progress (loss, accuracy) and inference results (class
predictions with confidence scores) are rendered in real time.
The system also
operates without OLED using serial output only.}
\label{fig:interface}
\end{figure}

Figure~\ref{fig:interface} shows the capacitive touch-based menu system.

\subsection{Class Configuration}
\label{subsec:classes}

The entire system is driven by two lines at the top of the sketch:

\begin{lstlisting}[caption={Configuring class count and names (Part~0).}]
#define NUM_CLASSES 3
String myClassLabels[NUM_CLASSES] = {"0Blank", "1Cup", "2Pen"};
\end{lstlisting}

Changing \texttt{NUM\_CLASSES} and updating the label array reconfigures the
network output layer, SD folder structure, OLED display, and inference printout
at compile time.
The reference dataset (blank page, child's sippy cup upside-down, ballpoint pen)
is included in \texttt{on-device-vision-ai} as a minimal proof-of-concept only.
An important practical consideration: as the number of classes grows, so does
the size of the dense output layer
(\texttt{NUM\_CLASSES} $\times$ \texttt{FLATTENED\_SIZE} weights) and the
potential for inter-class confusion.
Practitioners should expect to increase dataset size, training epochs, and
possibly network depth as class count increases, while monitoring available
PSRAM to remain within the 8~MB budget.

\subsection{Weight Priority System}
\label{subsec:weights}

Boot-time weight initialization is resolved automatically in \texttt{setup()}
via a three-tier priority:

\begin{enumerate}
  \item \textbf{SD binary} (\texttt{/header/myWeights.bin}): loaded first if
        present, enabling seamless resumption of training across power cycles.
  \item \textbf{Baked-in header} (\texttt{myWeights.h}, activated by
        \texttt{\#define USE\_BAKED\_WEIGHTS}): compiled into flash after
        copying the exported header from SD.
        If an SD binary is also present, it takes priority at the next boot.
  \item \textbf{Random He initialization}: used when no pre-trained weights
        are present.
\end{enumerate}

A key design point: when \texttt{USE\_BAKED\_WEIGHTS} is active, the SD card
is \emph{not} disabled.
Inserting an SD card and running additional training saves new weights as
\texttt{myWeights.bin}, which overrides the baked-in weights at the next boot
without recompilation.
This supports the full development arc: random init $\to$ on-device training
$\to$ baked-in deployment $\to$ field fine-tuning via SD override.

\subsection{Neural Network Architecture}
\label{subsec:nn}

\begin{table}[H]
\centering
\caption{CNN architecture for \texttt{INPUT\_SIZE = 64}.}
\label{tab:arch}
\begin{tabular}{llll}
\toprule
\textbf{Layer} & \textbf{Output shape} & \textbf{Parameters} & \textbf{Notes} \\
\midrule
Input          & 64$\times$64$\times$3  & ---     & RGB, normalized to $[0,1]$ \\
Conv1 (3$\times$3, 4f) & 62$\times$62$\times$4 & 108+4  & Leaky ReLU ($\alpha$=0.1) \\
MaxPool 2$\times$2     & 31$\times$31$\times$4 & ---    & Argmax retained for backprop \\
Conv2 (3$\times$3, 8f) & 29$\times$29$\times$8 & 288+8  & Leaky ReLU ($\alpha$=0.1) \\
Flatten        & 6{,}728               & ---     & \\
Dense          & \texttt{NUM\_CLASSES} & 20{,}184+3 & Softmax, Kahan summation \\
\midrule
\textbf{Total} & & \textbf{20{,}595} & \\
\bottomrule
\end{tabular}
\end{table}

The two convolutional layers are defined entirely through \texttt{\#define}
constants: kernel size and filter count for each layer.
Weight-buffer sizes and all intermediate tensor dimensions follow automatically,
so the depth and width of the network can be adjusted without touching the
forward- or backward-pass code:

\begin{lstlisting}[language=C++]
#define CONV1_KERNEL_SIZE 3
#define CONV1_FILTERS     4
#define CONV2_KERNEL_SIZE 3
#define CONV2_FILTERS     8
// Weight counts derived automatically:
#define CONV1_WEIGHTS  (CONV1_KERNEL_SIZE * CONV1_KERNEL_SIZE \
                        * 3 * CONV1_FILTERS)             //  108
#define CONV2_WEIGHTS  (CONV2_KERNEL_SIZE * CONV2_KERNEL_SIZE \
                        * CONV1_FILTERS * CONV2_FILTERS) //  288
#define OUTPUT_WEIGHTS (FLATTENED_SIZE * NUM_CLASSES)    // 20184
\end{lstlisting}

The single \texttt{INPUT\_SIZE} preprocessor constant scales all downstream
dimensions at compile time:

\begin{lstlisting}[language=C++]
#define INPUT_SIZE 64
// Derived automatically:
#define CONV1_OUTPUT_SIZE (INPUT_SIZE - 2)            // 62
#define POOL1_OUTPUT_SIZE (CONV1_OUTPUT_SIZE / 2)     // 31
#define CONV2_OUTPUT_SIZE (POOL1_OUTPUT_SIZE - 2)     // 29
#define FLATTENED_SIZE    (CONV2_OUTPUT_SIZE \
                           * CONV2_OUTPUT_SIZE \
                           * CONV2_FILTERS)           // 6728
\end{lstlisting}

\section{Implementation Details}
\label{sec:implementation}

\subsection{Memory Management}
\label{subsec:memory}

All large tensors---weights, gradients, Adam momentum buffers, activation
maps---are allocated in 8~MB PSRAM via \texttt{ps\_malloc()} and
\texttt{ps\_calloc()} in a single startup function (\texttt{myAllocateMemory()}).
Adam moment buffers are zero-initialized via \texttt{ps\_calloc()} to avoid
undefined state at the first optimizer step.
A 172~KB byte buffer (\texttt{myRgbBuffer}: \texttt{uint8\_t[240$\times$240$\times$3]})
is allocated once in \texttt{setup()} and reused across all inference frames,
training image loads, and OLED rendering, avoiding the repeated PSRAM
fragmentation and latency caused by per-image dynamic allocation.
Gradient buffers are allocated at program start alongside weights, keeping the
full memory footprint predictable.
If any critical allocation fails, the system halts with an OLED error message
rather than entering an undefined state.

Numerical stability is maintained through:
\begin{itemize}[nosep]
  \item \textbf{Kahan compensated summation} in the dense layer, eliminating
        floating-point drift across 6{,}728-term dot products.
  \item \textbf{Gradient clipping} to $[-100, 100]$ with NaN/Inf validation
        after each backward pass.
  \item \textbf{Weight clipping} to $[-10, 10]$ inside the Adam update,
        preventing weight explosion in early epochs.
  \item \textbf{Max-subtracted softmax} preventing overflow in exponential
        computation.
\end{itemize}

Memory footprint at \texttt{INPUT\_SIZE = 64}: full system (weights + gradients
+ Adam buffers + activations) $\sim$750~KB (9.4\% of PSRAM);
inference-only footprint (no gradient/Adam buffers) $\sim$235~KB (2.9\% of
PSRAM).

\subsection{Forward Propagation}
\label{subsec:forward}

Convolutional layers use explicit nested loops that map one-to-one onto the
standard definition of 2D convolution.
Filter-stride offsets (e.g., \texttt{ob = f * CONV1\_OUTPUT\_SIZE *
CONV1\_OUTPUT\_SIZE}) are hoisted outside the inner spatial loops to avoid
repeated multiplication on each pixel:

\begin{lstlisting}[caption={Conv1 forward pass with pre-computed offsets (Part~2).}]
for(int f = 0; f < CONV1_FILTERS; f++) {
  int ob = f * CONV1_OUTPUT_SIZE * CONV1_OUTPUT_SIZE;  // hoisted
  for(int y = 0; y < CONV1_OUTPUT_SIZE; y++) {
    for(int x = 0; x < CONV1_OUTPUT_SIZE; x++) {
      float sum = 0;
      for(int ky = 0; ky < 3; ky++) {
        for(int kx = 0; kx < 3; kx++) {
          int inPos = ((y+ky)*INPUT_SIZE + (x+kx)) * 3;
          int wPos  = f*27 + ky*9 + kx*3;
          sum += input[inPos  ] * myConv1_w[wPos  ]
               + input[inPos+1] * myConv1_w[wPos+1]
               + input[inPos+2] * myConv1_w[wPos+2];
        }
      }
      myConv1_output[ob + y*CONV1_OUTPUT_SIZE + x]
        = leaky_relu(clip_value(sum + myConv1_b[f]));
    }
  }
}
\end{lstlisting}

The verbosity is intentional: any intermediate activation can be observed live
in the Arduino serial monitor by adding a single \texttt{Serial.printf()} line,
without rebuilding a framework.

\subsection{Optimized Inference Resize}
\label{subsec:infresize}

Earlier firmware versions computed per-pixel floating-point division to map the
240$\times$240 camera frame to the 64$\times$64 input tensor on every inference
frame.
The current system pre-computes two integer lookup tables
(\texttt{sy\_lookup}, \texttt{sx\_lookup}) once on first entry to inference
mode, and replaces the per-pixel \texttt{/ 255.0f} normalization with a
pre-computed reciprocal multiplication (\texttt{* 0.003921569f}):

\begin{lstlisting}[caption={Pre-computed resize lookup tables with multiply-based normalization (Part~3).}]
static int  sy_lookup[INPUT_SIZE];
static int  sx_lookup[INPUT_SIZE];
static bool lookup_initialized = false;
if (!lookup_initialized) {
  for(int i = 0; i < INPUT_SIZE; i++) {
    sy_lookup[i] = min((int)((i+0.5f)*240.0f/INPUT_SIZE), 239);
    sx_lookup[i] = sy_lookup[i];   // square crop, same scale
  }
  lookup_initialized = true;
}
// Per-frame inner loop: two array lookups, no division
for(int y = 0; y < INPUT_SIZE; y++) {
  for(int x = 0; x < INPUT_SIZE; x++) {
    int srcIdx = (sy_lookup[y]*240 + sx_lookup[x]) * 3;
    int dstIdx = (y*INPUT_SIZE + x) * 3;
    myInputBuffer[dstIdx  ] = myRgbBuffer[srcIdx  ] * 0.003921569f;
    myInputBuffer[dstIdx+1] = myRgbBuffer[srcIdx+1] * 0.003921569f;
    myInputBuffer[dstIdx+2] = myRgbBuffer[srcIdx+2] * 0.003921569f;
  }
}
\end{lstlisting}

These optimizations reduce per-frame latency from $\sim$250~ms (4.0~FPS) to
$\sim$159~ms (\textbf{6.3~FPS}).

\subsection{Backpropagation and Batch Gradient Accumulation}
\label{subsec:backprop}

Weight gradient accumulators (\texttt{myConv1\_w\_grad},
\texttt{myConv2\_w\_grad}, \texttt{myOutput\_w\_grad}) are zeroed
\emph{once per batch}---before iterating over the images in that batch---so
that every image in the mini-batch contributes to the Adam update:

\begin{lstlisting}[caption={Batch-level gradient zeroing before Adam update (Part~2).}]
// Zero weight gradient accumulators once per batch.
memset(myConv1_w_grad,  0, CONV1_WEIGHTS  * sizeof(float));
memset(myConv1_b_grad,  0, CONV1_FILTERS  * sizeof(float));
memset(myConv2_w_grad,  0, CONV2_WEIGHTS  * sizeof(float));
memset(myConv2_b_grad,  0, CONV2_FILTERS  * sizeof(float));
memset(myOutput_w_grad, 0, OUTPUT_WEIGHTS * sizeof(float));
memset(myOutput_b_grad, 0, NUM_CLASSES    * sizeof(float));

for(int i = batchStart; i < batchEnd; i++) {
  // forward + backward; weight grads use +=, accumulating
}
myUpdateWeights(batch + 1);  // Adam step on fully accumulated grads
\end{lstlisting}

Per-image propagation buffers (\texttt{myDense\_grad}, \texttt{myPool1\_grad},
\texttt{myConv1\_grad}) are zeroed per-image, as they carry the error signal
\emph{through} the network rather than accumulating weight updates.
This two-level zeroing discipline---per-batch for weight accumulators,
per-image for propagation buffers---is essential for correct mini-batch
gradient descent and is documented explicitly in the source code.

\subsection{Adam Optimizer}
\label{subsec:adam}

Standard bias-corrected Adam~\cite{kingma2015adam}:

\begin{lstlisting}[caption={Adam update (Part~2).}]
float b1=0.9f, b2=0.999f, eps=1e-6f;
float lr_t = LEARNING_RATE * sqrtf(1.0f - powf(b2, step))
                           / (1.0f - powf(b1, step));
for(int i = 0; i < size; i++) {
  m[i] = b1*m[i] + (1-b1)*g[i];
  v[i] = b2*v[i] + (1-b2)*g[i]*g[i];
  w[i] -= lr_t * m[i] / (sqrtf(v[i]) + eps);
  w[i]  = clip_value(w[i], -10.0f, 10.0f);
}
\end{lstlisting}

Default hyperparameters: $\alpha=0.0003$, $\beta_1=0.9$, $\beta_2=0.999$,
$\epsilon=10^{-6}$ (raised from the conventional $10^{-8}$ to prevent NaN
in float32 at early training steps when second-moment estimates are near
zero), batch size~6, 20~epochs per run.
All are defined as mutable globals modifiable in the sketch.

\subsection{Configurable Validation Split}
\label{subsec:validation}

\texttt{VALIDATION\_IMAGES} (default: 3 per class) reserves the last $N$ images
in each class folder---sorted by filename for a deterministic split---as a
hold-out set.
After the final epoch, the system reports:

\begin{lstlisting}[language={}, caption={Post-training validation output.}]
--- Training Complete ---
Validation Accuracy: 77.8%  (7/9 correct)
\end{lstlisting}

This first generalization signal motivates further study of augmentation,
dropout, and cross-validation---topics that flow naturally in a course setting.

\subsection{Weight Export}
\label{subsec:export}

After training, the system writes two files to \texttt{/header/} on the SD card:

\begin{itemize}[nosep]
  \item \texttt{myWeights.bin}: raw 32-bit float arrays for rapid binary
        reload ($<$0.1~s at next boot).
  \item \texttt{myWeights.h}: a self-documenting C header with generation
        metadata, class label comments, and instructions for enabling
        \texttt{USE\_BAKED\_WEIGHTS}.
\end{itemize}

\begin{lstlisting}[caption={Fragment of exported myWeights.h.}]
// Auto-generated by on-device-vision-ai v1.0.0
//   #define NUM_CLASSES 3
//   String myClassLabels[] = {"0Blank", "1Cup", "2Pen"};
// To use: copy to sketch folder, then uncomment:
//   #define USE_BAKED_WEIGHTS
const float myModel_conv1_w[] = { /* 108 values */ };
// ... (all layers)
\end{lstlisting}

\subsection{Touch Interface and Responsive Exit}
\label{subsec:touch}

The multi-tap capacitive touch system (Pin~A0) uses configurable thresholds
(\texttt{myThresholdPress = 1100}, \texttt{myThresholdRelease = 900}) with
50~ms debounce and an 800~ms tap-window.
Three or more taps within the window are interpreted as a long-press (menu
confirm or exit action);
fewer taps register as a standard tap (menu advance or image capture).
Touch state is checked every third training image during backpropagation,
allowing the user to interrupt and save weights without waiting up to
15~seconds for a batch boundary.

\section{Experimental Evaluation}
\label{sec:evaluation}

\subsection{Dataset and Setup}
\label{subsec:dataset}

The experiments use a custom three-class dataset collected entirely on-device:
\texttt{0Blank} (blank white page), \texttt{1Cup} (child's sippy cup,
upside-down), and \texttt{2Pen} (ballpoint pen or pencil).
Images are captured at 240$\times$240 JPEG (with horizontal mirror for natural
orientation) and downsampled to 64$\times$64 RGB during training.
Three reference datasets of increasing size are released alongside the source
code: Dataset~A (2~classes~$\times$~9~images per class) as a smoke-test to verify
the installation runs end-to-end; Dataset~B (3~classes~$\times$~$\sim$30~images
per class, $\sim$90~images total, matching the experiments reported here) as a
balanced small-scale training set; and Dataset~C (6~classes~$\times$~$\sim$60~images)
for exploring class-count scaling.
The performance results reported here use Dataset~B, available directly at:\\
\url{https://github.com/webmcu-ai/on-device-vision-ai/tree/main/datasets/dataset-b}\\
All datasets are intentionally minimal and provided as proof-of-concept starting
points.
Larger datasets, including established benchmarks, produce proportionally
higher accuracy.

Training configuration: 64$\times$64 input, batch size~6, learning rate~0.0003,
20 epochs per run.
All configurable in the sketch.

\subsection{Training and Inference Performance}
\label{subsec:perf}

\begin{table}[H]
\centering
\caption{System performance on ESP32-S3 at \texttt{INPUT\_SIZE = 64}.}
\label{tab:perf}
\begin{tabular}{ll}
\toprule
\textbf{Metric} & \textbf{Value} \\
\midrule
Total parameters & 20{,}595 \\
Training memory (peak) & $\sim$750~KB (9.4\% of PSRAM) \\
Inference memory & $\sim$235~KB (2.9\% of PSRAM) \\
Training time (20 epochs, 90 images) & $\sim$9~min per run \\
\textbf{Inference rate (serial output)} & \textbf{6.3~FPS (159~ms/frame)} \\
Inference rate (OLED overlay, every 10th frame) & $\sim$0.8~FPS effective display \\
Compilation time (i9 laptop, 32~GB RAM) & $<$1~minute \\
Typical training accuracy (end of run) & $\sim$71--79\% \\
Typical validation accuracy & $\sim$67--78\%  \\
\bottomrule
\end{tabular}
\end{table}

Live inference serial output:

\begin{lstlisting}[language={}, caption={Live inference at 6.3~FPS.}]
Frame 3: 159 ms (6.3 FPS)
Pred: 0Blank (55.2%) | All: 55% 20% 25%
Frame 4: 159 ms (6.3 FPS)  Pred: 2Pen   (59.3%) |
All: 38%  2% 59%
Frame 5: 162 ms (...
\end{lstlisting}

The per-class probability breakdown gives students immediate intuition for
softmax output and confidence calibration.
Training convergence example:

\begin{lstlisting}[language={}, caption={Training output (serial monitor).}]
Batch 5/32 - Loss: 0.7215 - Acc: 66.7%
Batch 10/32 - Loss: 0.5803 - Acc: 71.4%
Batch 32/32 - Loss: 0.4831 - Acc: 78.6%
--- Training Complete ---
Validation Accuracy: 77.8% (7/9 correct)
\end{lstlisting}

\paragraph{Iterative training.}
The system supports iterative training: weights saved at the end of one run are
loaded automatically at the start of the next.
Empirically, meaningful accuracy improvements are observed across successive
runs on the reference dataset, with training accuracy approaching saturation
after two or three runs (total wall-clock time $\sim$18--27~minutes).
The system prints ``Continuing from saved weights'' or ``Starting fresh
training'' at the start of each session, making the iterative progression
explicit for learners.

\subsection{Energy Consumption}
\label{sec:energy}

\begin{figure}[t]
\centering
\includegraphics[width=\textwidth]{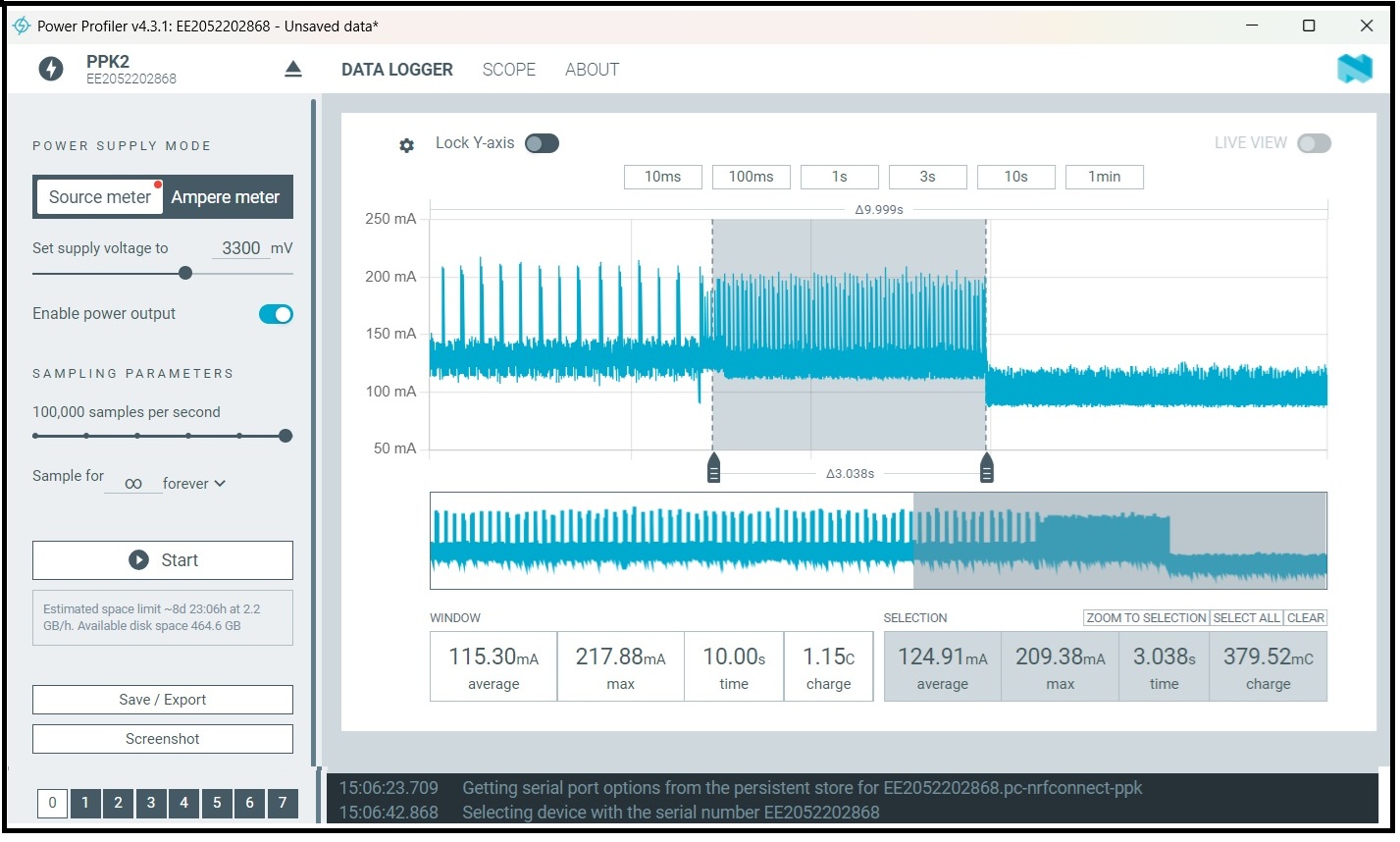}
\caption{Power profile of an on-device training cycle at 3.3~V, measured with a
Nordic PPK2.}
\label{fig:power}
\end{figure}

Power was measured at 3.3~V using a Nordic PPK2 power profiler
(Figure~\ref{fig:power}).

\begin{table}[H]
\centering
\caption{Power profile at 3.3~V.}
\label{tab:power}
\begin{tabular}{lll}
\toprule
\textbf{Mode} & \textbf{Current} & \textbf{Power} \\
\midrule
Idle & 100--120~mA & 0.33--0.40~W \\
Image collection & 130--150~mA & 0.43--0.50~W \\
Training (compute) & 150--180~mA & 0.50--0.59~W \\
Training (SD write) & 180--220~mA & 0.59--0.73~W \\
Inference (6.3~FPS) & 120--150~mA & 0.40--0.50~W \\
Peak (SD ops) & 220~mA & 0.73~W \\
\midrule
Energy per training run & \multicolumn{2}{l}{$\sim$0.07~Wh (20 epochs, 90 images, $\sim$9~min)} \\
\bottomrule
\end{tabular}
\end{table}

A selected 3.038~s training segment consumed 379.52~mC (124.91~mA average).
Periodic 200~mA spikes correspond to backpropagation through the dense layer;
SD write bursts produce the highest instantaneous current.

To the author's knowledge, few embedded ML training systems report end-to-end energy
consumption at this level of granularity, particularly for full training cycles
rather than inference alone.
This transparency enables direct comparison between computation and data
transmission costs in resource-constrained deployments.

The power trace (Figure~\ref{fig:power}) reveals distinct periodic current
spikes reaching approximately 200~mA, corresponding to compute-intensive phases
of training, particularly backpropagation through the dense layer.
Between these spikes, the system maintains a relatively stable baseline current
of approximately 110--130~mA.
This burst-pattern behavior indicates that training load is not uniformly
distributed over time, which has implications for power supply design, battery
sizing, and duty-cycled operation in embedded deployments.

Unlike typical ML system evaluations that report only average power, this
time-resolved measurement exposes the internal structure of the training
process, allowing direct correlation between algorithmic phases and energy
consumption.

\section{Scalability}
\label{sec:scalability}

Table~\ref{tab:scaling} gives approximate scaling behavior as \texttt{INPUT\_SIZE}
is varied.
All downstream dimensions---convolutional output sizes, pooling outputs, flattened
size, and dense layer weight count---are derived from \texttt{INPUT\_SIZE} at
compile time, so changing a single constant rescales the entire network.

\begin{table}[H]
\centering
\caption{Approximate scaling with \texttt{INPUT\_SIZE}.}
\label{tab:scaling}
\begin{tabular}{lllll}
\toprule
\textbf{INPUT\_SIZE} & \textbf{Parameters} & \textbf{Training memory} & \textbf{Train time/run} & \textbf{Inference} \\
\midrule
48  & $\sim$11{,}000 & $\sim$400~KB & $<$6~min  & $>$8~FPS \\
64  & 20{,}595       & $\sim$750~KB & $\sim$9~min & 6.3~FPS \\
96  & $\sim$46{,}000 & $\sim$1.7~MB & $\sim$21~min & $\sim$3~FPS \\
128 & $\sim$82{,}000 & $\sim$3.0~MB & $\sim$36~min & $\sim$2~FPS \\
\bottomrule
\end{tabular}
\end{table}

\paragraph{Scope of evaluation.}
All results are demonstrated on a single three-class vision classification task
using a custom on-device dataset.
While sufficient to validate system functionality and training behavior, broader
generalization across datasets, domains, and larger class counts remains future
work.

\section{Discussion}
\label{sec:discussion2}

\subsection{Comparison to Relevant Frameworks}
\label{subsec:comparison}

\begin{table}[H]
\centering
\caption{Comparison of ML systems for ESP32-S3.}
\label{tab:comparison}
\begin{tabular}{p{3.8cm} p{1.8cm} p{1.8cm} p{1.4cm} p{1.4cm}}
\toprule
\textbf{Aspect} & \textbf{This work} & \textbf{ESP-DL/ESP-NN} & \textbf{AIfES} & \textbf{MIT TTE} \\
\midrule
On-device training    & Yes  & No          & Yes     & Yes \\
On-device inference   & Yes  & Yes         & Yes     & Yes \\
INT8 quantization     & No   & Yes         & Partial & Yes \\
Code transparency     & Full & Proprietary & Partial & Limited \\
Energy measurability  & Full & Inference only & Partial & Partial \\
External toolchain    & None & Required    & Optional & Required \\
Single-file Arduino   & Yes  & No          & No      & No \\
Production ready      & Educational & Yes      & Partial & No \\
\bottomrule
\end{tabular}
\end{table}

The primary distinction is purpose.
ESP-DL and ESP-NN excel at executing a pre-trained, pre-quantized production
model on ESP32-S3 hardware; they are the correct choice for deployment at scale.
This work is the correct choice when training itself must occur on the device,
or when a practitioner needs to understand what training is doing before trusting
a framework to do it invisibly.
A complete ML engineering curriculum should expose students to both.

\paragraph{Reference to off-device training.}
For context, equivalent models trained off-device using standard frameworks
(e.g., TensorFlow or PyTorch) on similar datasets typically achieve higher
accuracy and faster convergence.
However, such workflows do not expose the training process, nor do they operate
under the same memory, power, and transparency constraints as a fully on-device
system.

\subsection{Design Decisions}
\label{subsec:design}

\paragraph{Float32.}
The ESP32-S3 hardware FPU makes 32-bit float natural and stable.
INT8 quantization is a natural follow-on experiment for students who have
understood the full-precision baseline.

\paragraph{Leaky ReLU.}
Prevents dying neurons without requiring additional debugging tools.
Swapping in ReLU, ELU, or GELU requires changing one function definition.

\paragraph{Explicit loops.}
Every convolution and pooling layer uses nested \texttt{for} loops rather than
SIMD intrinsics: slower, but fully traceable.
Any activation can be observed live in the serial monitor with one added
\texttt{Serial.printf()}.

\paragraph{Horizontal mirror.}
Camera output is mirrored horizontally (\texttt{set\_hmirror(s,~1)}) for a
natural front-facing orientation, reducing labeling errors during data
collection.

\paragraph{Single file.}
All $\sim$1,750 lines in one \texttt{.ino} file.
Unconventional for production code, but ideal for education: the entire system
can be read and forked in the Arduino IDE without navigating a multi-directory
project.

\subsection{Limitations}
\label{subsec:limitations}

\begin{itemize}[nosep]
  \item Single-core; the second Xtensa core is unused and available for
        parallelization.
  \item No data augmentation; accuracy is limited by dataset size and on-device
        collection conditions.
  \item No INT8 quantization; inference speed is below ESP-DL production
        performance.
  \item Platform-specific, though the principles and code structure transfer to
        any board with $>$4~MB external RAM (e.g., Arduino Portenta H7 with
        Vision Shield).
  \item The reference datasets are intentionally minimal.
        Reported accuracies are on training data and a small held-out set;
        overfitting cannot be ruled out at these dataset sizes.
  \item The system is not production-ready for high-volume deployment, though it
        is suitable for small business and prototype use cases.
\end{itemize}

\subsection{Alignment with webmcu-ai Series Goals}
\label{subsec:seriesgoals}

This paper documents a complete, self-contained on-device ML system and forms
the shared firmware foundation for the series of papers that follow.
The planned series covers four modalities: vision (this paper and Paper~2),
audio (Paper~3), and motion (Paper~4).
Each paper pair follows the same structure: a single-file Arduino
\texttt{.ino} for fully on-device operation, and a companion web application
connected via the WebSerial API.
Paper~2 will extend this vision system with a browser interface; a link will
be added to this paper upon publication.
This paper is designed to stand alone: all contributions described here are
fully realized within the single-file firmware and the reference datasets
released in \texttt{on-device-vision-ai}.

\subsection{Context and Accessibility}
\label{subsec:context}

This system was developed using commodity hardware and freely available tools,
without reliance on institutional research infrastructure or external funding.
The complete pipeline---data collection, training, evaluation, and
deployment---runs on a low-cost microcontroller platform, with core hardware
available at approximately \$15--40~USD.

This constraint is intentional rather than incidental.
It demonstrates that meaningful machine learning systems---encompassing gradient
correctness, optimizer behavior, energy measurement, and deployment---can be
conducted in resource-constrained environments while remaining transparent and
reproducible.

The resulting system is both a research artifact and a directly deployable
educational platform, lowering barriers for students, educators, and independent
practitioners to explore end-to-end ML systems without dependence on cloud
infrastructure.

\section{Future Work}
\label{sec:future}

\subsection{What You Can Do With This System Right Now}
\label{subsec:what_you_can_do}

The primary goal of this paper is to establish a transparent, working baseline
that others can build on immediately.
The firmware is a single Arduino sketch with no external ML dependencies: fork
it, read every line, change two lines to reconfigure the class labels, and train
your own classifier in under ten minutes of code modification.

Concrete starting points include:

\begin{itemize}[nosep]
  \item \textbf{New datasets.} Point the camera at any visual
        distinction---product defects, plant conditions, bin fill levels,
        attendance tokens---and collect images on-device.
        Dataset~A (2~classes, 9~images per class) verifies the pipeline runs
        end-to-end; Dataset~B (3~classes, $\sim$30~images per class) is the
        reference training set; Dataset~C (6~classes, $\sim$60~images per class)
        demonstrates class-count scaling.
  \item \textbf{More classes.} Change \texttt{NUM\_CLASSES} and add labels.
        The network output layer, SD folder structure, OLED display, and
        inference printout all reconfigure automatically.
        Monitor PSRAM headroom and expect to increase dataset size
        proportionally.
  \item \textbf{Larger input.} Change \texttt{INPUT\_SIZE} from 64 to 96 or
        128; all downstream dimensions derive from that constant.
        Table~\ref{tab:scaling} gives timing and memory estimates.
  \item \textbf{Algorithm experiments.} Swap Leaky ReLU for ReLU, ELU, or
        GELU by changing one function definition.
        Adjust batch size, learning rate, or epsilon in the Adam update.
        Add a \texttt{Serial.printf()} anywhere to observe live gradients or
        activations without rebuilding a framework.
  \item \textbf{Energy-aware deployment.} Use the power profiling results
        (Table~\ref{tab:power}) as a baseline; instrument your own dataset
        collection and training cycle to compare energy per class.
  \item \textbf{Field refinement.} Deploy with \texttt{USE\_BAKED\_WEIGHTS}
        for SD-free inference, then insert an SD card later to fine-tune on
        new images without recompilation.
\end{itemize}

All source code and reference datasets are at
\url{https://github.com/webmcu-ai/on-device-vision-ai}.

\subsection{Planned Extensions}
\label{subsec:planned}

Three follow-on papers are planned in this series.
Paper~2 will document the companion browser-based WebSerial interface
(\texttt{webmcu-vision-web}), which connects the same ESP32-S3 firmware to a
Chromium browser with no software installation: SD file management, live loss
and confusion matrix charts, weight download, and a JSON configuration file for
adjusting hyperparameters without recompilation.
Paper~3 will extend the framework to on-device audio classification using the
XIAO ML Kit's built-in I2S microphone.
Paper~4 will extend it to IMU-based gesture and activity recognition.
Each paper will follow the same two-repository structure: a single
\texttt{.ino} for fully on-device operation and a companion WebSerial web
application.

Community contributions and longer-term directions---including INT8 quantization
compatible with ESP-NN, dual-core parallelization, data augmentation, dropout,
FOMO object detection, and multi-device federated learning---are tracked as open
issues at \url{https://github.com/webmcu-ai} under the MIT License.

\section{Deployment Workflow}
\label{sec:deployment}

Complete step-by-step deployment instructions for both the Arduino IDE and
PlatformIO are maintained in the repository README at
\url{https://github.com/webmcu-ai/on-device-vision-ai}.
This includes board configuration, PSRAM settings, SD card preparation, serial
monitor navigation, and the baked-in weight workflow.
Keeping the workflow in the repository ensures it stays current as the firmware
evolves.

\section{Conclusion}
\label{sec:conclusion}

This work demonstrates that complete CNN training---data capture,
backpropagation, Adam optimization, configurable hold-out validation, and
real-time inference at 6.3~FPS---is achievable on a thumb-sized,
\$15--40~USD microcontroller.
The system implements correct batch-level gradient accumulation, PSRAM-aware
memory management with a globally pre-allocated RGB working buffer,
pre-computed inference lookup tables, multiply-based pixel normalization,
dual-format weight export, and a three-tier weight priority system supporting
the full development arc from initial training through baked-in deployment and
SD-based field refinement.

Where Espressif's own ESP-DL and ESP-NN are well-suited for deploying
a pre-trained model at scale, this system occupies a complementary position:
to the author's knowledge, the only documented option in which the core ML
pipeline---data collection, CNN training, and inference---executes on the
device itself, with no external computation.
The system is equally suited to small business and prototype deployments where
cloud dependency, ongoing cost, or data-privacy constraints make fully on-device
operation preferable.

To the author's knowledge, no prior work demonstrates a fully transparent, single-file
implementation of end-to-end CNN training and inference on the ESP32-S3 without
external computation.
This positions the system as both a functional ML platform and a pedagogical
tool for understanding embedded learning systems at the level of individual
operations.

The primary contribution is transparency: every gradient, every Adam moment,
every milliamp of training energy, and every line of the inference path are
directly observable and modifiable.
This is the property that production cloud frameworks cannot offer, and the
property that engineering education requires.

The system is designed for reproducibility: all source code is contained in a
single Arduino sketch with no external ML dependencies, datasets are provided,
and validation splits are deterministic.
This enables direct replication and modification by other practitioners without
specialized tooling.

All source code, the reference datasets, and this paper are available at
\url{https://github.com/webmcu-ai/on-device-vision-ai} under the MIT License.

\section*{Acknowledgements}

The author thanks Brian Plancher (Dartmouth College) for his generous support and
suggestions during this project; his contributions to the TinyML4D community~\cite{plancher2024tinyml4d}
provided both technical grounding and useful context for the on-device system described here.

The author used LLM assistants---Claude (Anthropic), ChatGPT (OpenAI),
Gemini (Google), and Copilot (Microsoft)---for structural editing and code
review; all technical decisions and results are the author's own.



\begin{thebibliography}{99}

\bibitem{espdl_github}
Espressif Systems, ``ESP-DL: Espressif Deep Learning Library,'' GitHub, 2024.
\url{https://github.com/espressif/esp-dl}

\bibitem{espnn_github}
Espressif Systems, ``ESP-NN: Optimized Neural Network Functions for ESP SoCs,'' GitHub, 2024.
\url{https://github.com/espressif/esp-nn}

\bibitem{reddi2024mlsysbook}
Vijay Janapa Reddi, ``MLSysBook.AI: Principles and Practices of Machine Learning Systems Engineering,'' Online textbook, 2024.
\url{https://mlsysbook.ai}

\bibitem{reddi2024tinytorch}
Vijay Janapa Reddi, ``TinyTorch: Building Machine Learning Systems from First Principles,'' arXiv:2601.19107 [cs.LG], January 2026.
\url{https://arxiv.org/abs/2601.19107}

\bibitem{lin2022ondevice}
J.~Lin, L.~Zhu, W.-M. Chen, W.-C. Wang, C.~Gan, and S.~Han, ``On-Device Training Under 256KB Memory,'' \textit{Advances in Neural Information Processing Systems (NeurIPS)}, vol.\ 35, pp.\ 13351--13361, 2022.
\url{https://arxiv.org/abs/2206.15472}

\bibitem{wulfert2024aifes}
L.~Wulfert et~al., ``AIfES: A Next-Generation Edge AI Framework,'' \textit{IEEE Trans.\ Pattern Anal.\ Mach.\ Intell.}, vol.\ 46, no.\ 6, pp.\ 4519--4533, June 2024.
\url{https://doi.org/10.1109/TPAMI.2024.3355495}

\bibitem{aifes_github}
Fraunhofer IMS, ``AIfES for Arduino,'' GitHub, 2024.
\url{https://github.com/Fraunhofer-IMS/AIfES_for_Arduino}

\bibitem{plancher2024tinyml4d}
B.~Plancher, S.~B\"{u}ttrich, J.~Ellis, et~al.,
``TinyML4D: Scaling Embedded Machine Learning Education in the Developing World,''
\textit{Proc.\ AAAI Symposium Series}, pp.\ 508--515, 2024.
\url{https://doi.org/10.1609/aaaiss.v3i1.31265}

\bibitem{kingma2015adam}
D.~P. Kingma and J.~Ba, ``Adam: A Method for Stochastic Optimization,'' \textit{Proc.\ Int.\ Conf.\ Learning Representations (ICLR)}, 2015.
\url{https://arxiv.org/abs/1412.6980}

\bibitem{ren2021tinyol}
H.~Ren et~al., ``TinyOL: TinyML with Online-Learning on Microcontrollers,'' arXiv preprint arXiv:2103.08295, 2021.
\url{https://arxiv.org/abs/2103.08295}

\bibitem{lin2024tinyprogress}
J.~Lin, L.~Zhu, W.-M. Chen, W.-C. Wang, and S.~Han, ``Tiny Machine Learning: Progress and Futures,'' arXiv preprint arXiv:2403.19076, 2024.
\url{https://arxiv.org/abs/2403.19076}

\bibitem{ren2024ondevice}
H.~Ren, D.~Anicic, and T.~A. Runkler, ``On-device Online Learning and Semantic Management of TinyML Systems,'' arXiv preprint arXiv:2405.07601, 2024.
\url{https://arxiv.org/abs/2405.07601}

\bibitem{karic2025send}
B.~Kari\'{c}, N.~Herrmann, J.~Stenkamp, P.~Scharf, F.~Gieseke, and A.~Schwering, ``Send Less, Save More: Energy-Efficiency Benchmark of Embedded CNN Inference vs.\ Data Transmission in IoT,'' arXiv preprint arXiv:2510.24829, 2025.
\url{https://arxiv.org/abs/2510.24829}

\bibitem{webserial2024}
W3C Web Incubator Community Group, ``Web Serial API,'' 2024.
\url{https://wicg.github.io/serial/}

\end{thebibliography}
\end{document}